\documentclass[runningheads]{llncs}

\usepackage[T1]{fontenc}
\usepackage{graphicx}
\usepackage{amsmath,amsfonts,amssymb,bm}
\usepackage{booktabs}
\usepackage{subcaption}
\usepackage{xcolor}
\usepackage[numbers,sort&compress]{natbib}
\usepackage{hyperref}

\begin{document}

\title{Beyond Random Partitioning: Unsupervised Spatio-Temporal
Stratification for Cohort Balancing in Longitudinal Medical Imaging}
\titlerunning{Spatio-Temporal Stratification for Cohort Balancing}


\author{Qinghui Liu\inst{1} \and Jon Andr\'{e} Ottesen\inst{1} \and Atle Bj\o{}rnerud\inst{1} \and Kyrre Eeg Emblem\inst{1}}
\authorrunning{Q. Liu et al.}
\institute{Department of Physics and Computational Radiology, Oslo University Hospital, Norway}


\maketitle

\begin{abstract}
Rigorous dataset partitioning is a foundational, yet frequently overlooked, prerequisite for reliable deep learning in longitudinal medical imaging. Naively shuffling small clinical cohorts routinely introduces covariate shifts and temporal sampling imbalances across training, validation, and test subsets, exposing downstream models to out-of-distribution evaluation. We address this vulnerability with an auditable Tripartite Dataset Analytics Framework that systematically characterizes spatial grid integrity, multi-parametric intensity fingerprints, and longitudinal temporal trajectories, quantifying the heavy-tailed feature dispersion and irregular, episodic sampling intervals typical of real-world clinical cohorts. Building on this characterization, we formalize an unsupervised spatio-temporal cohort-balancing standard operating procedure (SOP) that combines elbow-optimized $K$-means clustering over a standardized, six-dimensional joint intensity-temporal feature space with intra-cluster proportionate stratified sampling. On a longitudinal, contrast-enhanced $T1$-weighted brain MRI cohort ($N=149$), the protocol reduces the maximum cross-subset intensity bias from 34.1\% under conventional random shuffling to under 2.1\%, while aligning longitudinal follow-up intervals closely around the population mean. Monte Carlo stress testing across ten random seeds and three split configurations confirms that this alignment remains tightly bounded, in clear contrast to the substantial variability of random partitioning. The resulting protocol offers a reproducible, generalizable procedure for cohort engineering in variable-length longitudinal clinical imaging workflows.

\keywords{Cohort partitioning \and Longitudinal medical imaging \and Stratified sampling \and Domain shift.}

\end{abstract}

\section{Introduction}\label{sec:intro}

Clinical imaging follow-up is inherently irregular: encounters are episodic, and their timing and frequency vary widely across patients due to scheduling constraints and adherence. Treating these variable-length trajectories as exchangeable, static instances conflicts with the distributional assumptions underlying naive random partitioning. Nevertheless, medical image computing routinely relies on an unexamined ``shuffle-and-train'' paradigm, or on standard label-blind $K$-fold cross-validation, to split dataset cohorts\cite{zhang2019survey} -- a convention inherited from static computer-vision benchmarks. Applied to continuous temporal patient chains, this practice stochastically introduces artificial shifts between training and evaluation boundaries, pushing downstream deep architectures toward out-of-distribution evaluation failure\cite{tak2025longitudinal,disch2025temporal}.

\subsection{Problem Statement}

The methodological challenges of optimization under small-sample constraints in medical deep learning have been extensively documented. Zhang et al.\cite{zhang2019survey} provided a comprehensive taxonomy of algorithmic strategies targeting the small-sample-size problem in medical image analysis, partitioning solutions into advanced data augmentation, domain-specific transfer learning, and specialized architectural regularizers. Cheplygina et al.\cite{cheplygina2019not} further examined how weak-supervision, semi-supervised clustering, and multi-instance learning can leverage unannotated or imperfectly structured imaging repositories to reinforce representation stability in data-limited scenarios. However, these downstream architectural and algorithmic solutions typically presuppose that the underlying training, validation, and testing partitions maintain strict distributional equivalence---an essential assumption that is frequently violated in real-world clinical practice\cite{matta2024systematic}.

Recent systematic reviews have highlighted that domain shift constitutes a primary source of performance degradation during model deployment. Matta et al.\cite{matta2024systematic} analyzed 77 distinct clinical cross-scanner studies and concluded that cross-population domain shifts commonly degrade predictive fidelity by 10--25\% when models are evaluated on unseen populations. Yoon et al.\cite{yoon2024domain} further demonstrated that these shifts arise from demographic variability, scanner hardware variations, and center-specific imaging protocols, creating systemic covariate shifts that severely undermine model generalization bounds\cite{quinonero2009dataset}. This vulnerability is significantly compounded in longitudinal studies, where temporal sampling heterogeneity, variable follow-up durations, and irregular scanning sequences introduce complex multi-dimensional distribution skews that conventional unstratified shuffling cannot regularize\cite{tak2025longitudinal}.

\subsection{Failure Modes of Random Partitioning}

\begin{figure}[!tbp]
    \centering
    \includegraphics[width=\textwidth]{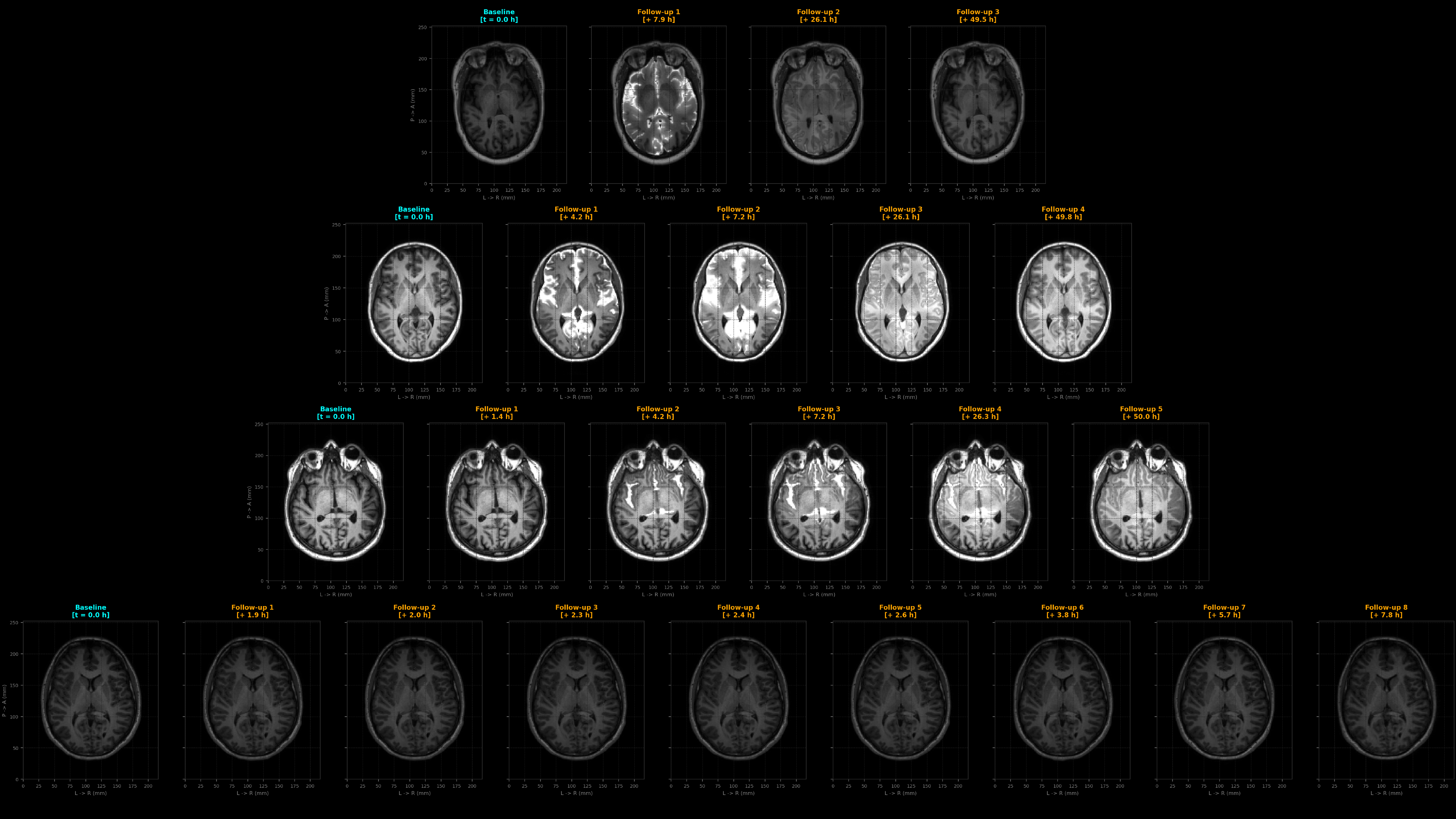}
    \caption{\textbf{Clinical motivation: representative longitudinal tracking timelines illustrating spatial contrast and temporal sampling heterogeneity.} Each row shows one patient tracked from the pre-contrast baseline scan ($t=0$~h) through subsequent follow-up encounters, rendered under a fixed contrast window ($[-0.337, 1]$) so that intensity differences reflect genuine pathology rather than display normalization. Rows 2 and 3 exhibit pronounced tracer accumulation and signal saturation ($I_{max} > 2.0$), while Rows 1 and 4 remain comparatively subdued; follow-up length ranges from $M=3$ to $M=8$ encounters across patients, illustrating the joint intensity--temporal heterogeneity that conventional random partitioning fails to account for.}
    \label{fig:longitudinal_patient_flows}
\end{figure}
We identify two primary statistical failure modes inherent to conventional random splitting when operating on variable-length longitudinal medical imaging cohorts:
\begin{enumerate}
    \item \textbf{Sampling Bias and Covariate Shift:} Multi-parametric medical imaging features typically exhibit high inter-subject variance, multi-modal topologies, and heavy-tailed density distributions driven by localized pathology or scanner-specific hardware parameters\cite{gao2019universal}. Unstratified splitting risks stochastically concentrating rare clinical sub-phenotypes, distinct intensity distributions, or statistical outliers within a single partition. This misallocation under-represents such cases in training while exposing the model to out-of-distribution conditions at evaluation, effectively inducing a covariate shift between partitions\cite{matta2024systematic,ghazvanchahi2024effect}. The intensity-normalization literature provides direct evidence for this vulnerability: Ghazvanchahi et al.\cite{ghazvanchahi2024effect} showed that deep learning networks evaluated on out-of-distribution data from mismatched scanners suffer substantial performance degradation, with regional similarity metrics dropping by up to 8\% without proper intensity standardization.

    \item \textbf{Temporal Sampling Imbalance and Structural Skewness:} In longitudinal clinical studies, follow-up frequency, tracking interval, and sequence length per subject are inherently non-uniform, driven by clinical scheduling, patient compliance, and operational constraints\cite{tak2025longitudinal,disch2025temporal}. Pure random allocation does not regulate the total number of derived sequential pairs across subsets, and frequently produces substantial imbalances in temporal sampling density across validation boundaries. Tak et al.\cite{tak2025longitudinal} showed that while sequential deep learning models improve longitudinal prediction stability by up to 58.5\% relative to single time-point analyses, differences in follow-up trajectories across subsets can substantially confound cross-validation metrics.

\end{enumerate}

Figure~\ref{fig:longitudinal_patient_flows} illustrates this variance across four representative clinical cases, rendered under a fixed intensity window ($[-0.337, 1]$) so that the visual contrast reflects genuine biological and pathological heterogeneity rather than display normalization. Cases with rapid contrast-tracer accumulation or hyperintense lesion boundaries (Rows 2--3) show pronounced, localized signal saturation ($I_{max} > 2.0$), while others remain visually subdued (Rows 1 and 4); follow-up length also varies widely, from $M=3$ to $M=8$ encounters. A random split can, by chance, assign most of the saturated, densely-sampled trajectories (e.g., Row 3, $M=5$) to one subset while concentrating sparse, low-contrast timelines (e.g., Row 1, $M=3$) in another, violating the cross-subset exchangeability assumption and introducing an artificial covariate shift between training and evaluation.
\subsection{Research Question \& Contributions}

Guided by these vulnerabilities, this study addresses a central methodological question: \textit{how can we construct verifiably homogeneous and stable dataset partitions for irregular, variable-length longitudinal clinical cohorts, without introducing manual bias, data leakage, or statistical confounding across evaluation splits?} To move from ad hoc shuffling toward reproducible, evidence-based cohort design, we formalize a generalized standard operating procedure (SOP) for clinical pre-training cohort engineering. Our contributions are threefold:
\begin{enumerate}
    \item We propose a \textbf{Tripartite Dataset Analytics Framework} that systematically audits spatial grid integrity, statistical multi-parametric intensity fingerprints, and continuous longitudinal trajectories prior to model training.
    \item We formalize a \textbf{Clustering-Based Stratified Sampling Algorithm} that maps heterogeneous subjects onto an isotropic, six-dimensional joint space, utilizing elbow-optimized $K$-means clustering and intra-cluster proportionate stratified sampling to enforce statistical homogeneity.
    \item We show empirically that conventional random splitting induces measurable, substantial covariate shifts in longitudinal clinical cohorts, and that our proposed SOP tightly restricts cross-subset variance across both holdout and iterative cross-validation regimes.
\end{enumerate}

\section{The Tripartite Dataset Analytics Framework}\label{sec:tripartite}

To make pre-training data curation auditable rather than opaque, we formalize a Tripartite Dataset Analytics Framework. Rather than treating cohort preparation as an administrative step, this framework represents the spatial, contrast, and temporal characteristics of a sequential medical dataset as structured statistical distributions, yielding an interpretable data pipeline that characterizes structural and statistical homogeneity prior to model training.

\subsection{Pillar 1: Spatial Grid and Metadata Integrity Auditing}\label{sec:pillar1}

Multi-dimensional deep learning architectures (e.g., 3D convolutional networks\cite{tak2025longitudinal}) rely on the assumption that inputs share a unified, isotropic spatial coordinate system. In heterogeneous or multi-center clinical cohorts, variations in acquisition protocols can introduce geometric misalignments, grid corruptions, or anisotropy anomalies\cite{ghazvanchahi2024effect,nyul1999standardizing}.

To address this, the first pillar establishes a spatial grid verification protocol. Let each volumetric image be defined by its discrete matrix $\mathbf{V} \in \mathbb{R}^{N_x \times N_y \times N_z}$, where $N_c$ denotes the grid resolution along cardinal axis $c \in \{X, Y, Z\}$. The absolute physical voxel dimension is parameterized by the spacing vector $\bm{\lambda} = [\lambda_x, \lambda_y, \lambda_z]$. The auditing protocol enforces a uniform spatial constraint across the cohort $\mathcal{D}$:
\begin{equation}\label{eq:spatial}
    \forall i, j \in \mathcal{D}, \quad [N_x^{(i)}, N_y^{(i)}, N_z^{(i)}] \equiv [N_x^{(j)}, N_y^{(j)}, N_z^{(j)}] \land \bm{\lambda}_i \equiv \bm{\lambda}_j
\end{equation}

\textit{Empirical Validation Case Study:} This auditing procedure was applied to our longitudinal structural MRI cohort ($T1$-weighted volumes), together with their corresponding binary anatomical masks and derived signed distance fields (SDFs). These raw clinical volumes first underwent a standard preprocessing pipeline: automated brain extraction using HD-BET\cite{isensee2019automated}, followed by isotropic resampling to $1.5\,\text{mm} \times 1.5\,\text{mm} \times 1.5\,\text{mm}$. To satisfy the strided pooling and convolution requirements of standard hierarchical architectures (e.g., U-Net encoder-decoders), the cropped volumes were zero-padded to dimensions divisible by 8. Pillar 1 then performed a file-by-file verification, confirming that 100\% of the retained sessions converged to a unified spatial grid of $144 \times 152 \times 168$ voxels -- ensuring that downstream convolutions operate on a geometrically consistent domain, preventing dimensional-mismatch failures prior to model execution.

\subsection{Pillar 2: Multi-Parametric Statistical Intensity Fingerprinting}\label{sec:pillar2}

Voxel-level intensity distributions are highly sensitive to interpersonal biological variation, scanner hardware drift, and localized tissue pathology\cite{ghazvanchahi2024effect,nyul1999standardizing,shinohara2014standardized}. The MRI intensity standardization literature has demonstrated that hardware and software variations create non-standard intensities, contrast distributions, and noise profiles across scanners, significantly impacting model generalization\cite{ghazvanchahi2024effect}. Unprofiled intensity distributions can also distort gradient backpropagation in deep architectures\cite{gao2019universal}.

\begin{figure}[!tbp]
    \centering
    \includegraphics[width=\textwidth]{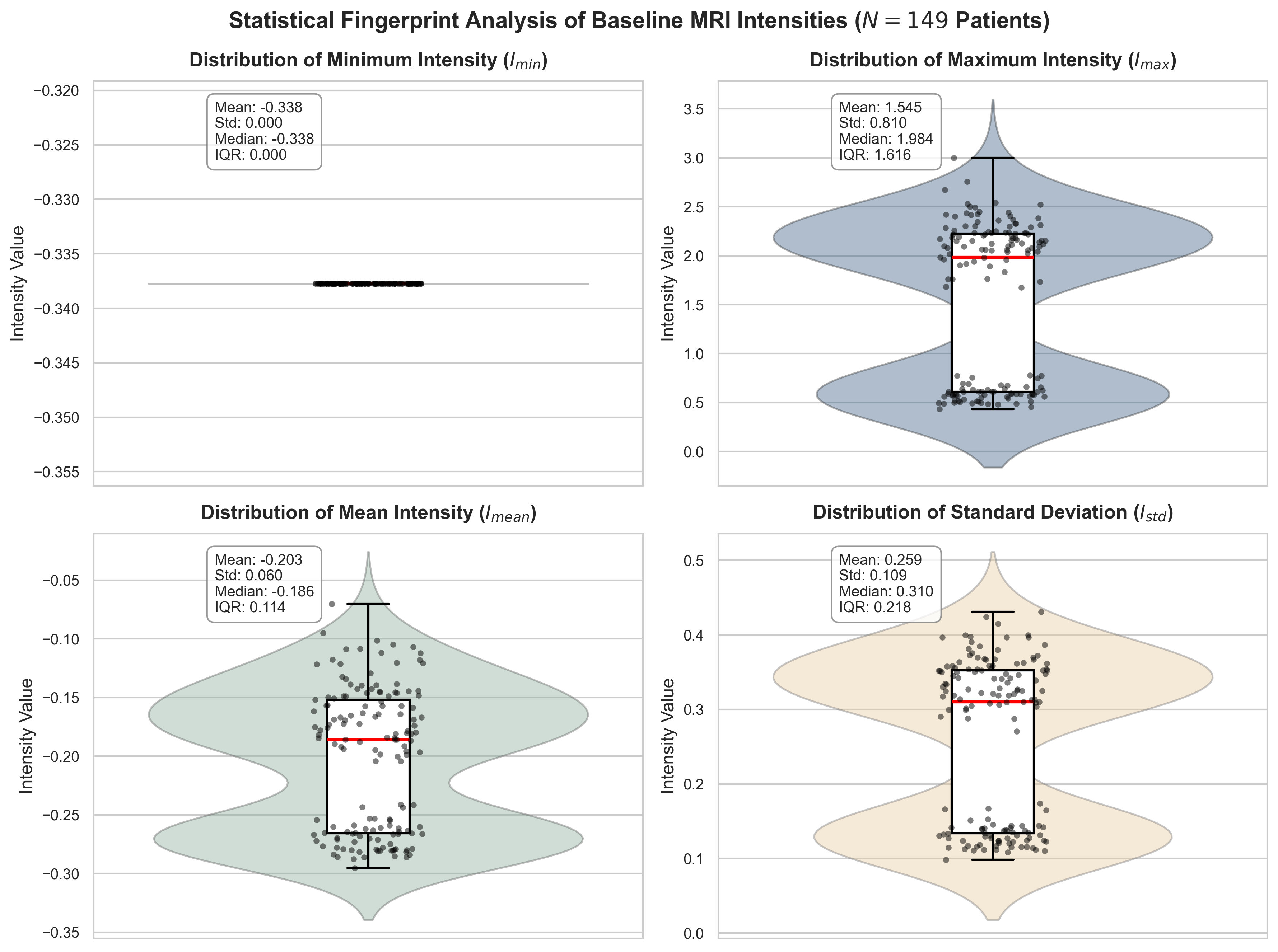}
    \caption{\textbf{Multi-parametric intensity fingerprint analysis of the baseline cohort.} The multi-panel plot combines continuous kernel density estimations (violin contours), quartile boundaries (boxplots with median demarcations), and jittered individual subject points, highlighting the sharp distributional contrast between the convergent $I_{min}$ values and the heavy-tailed $I_{max}$ parameter.}
    \label{fig:intensity_distributions}
\end{figure}

To characterize the macro-level intensity state of the cohort, this pillar extracts a four-dimensional statistical fingerprint $\bm{\chi}_i$ from the bounded anatomical volume of each subject $i$:
\begin{equation}\label{eq:fingerprint}
    \bm{\chi}_i = \left[ I_{min}^{(i)}, \, I_{max}^{(i)}, \, I_{mean}^{(i)}, \, I_{std}^{(i)} \right]
\end{equation}
where $I_{min}$ and $I_{max}$ define the absolute intensity bounds, while $I_{mean}$ and $I_{std}$ capture the first and second central moments of the voxel intensity distribution.

\begin{figure}[!tbp]
    \centering
    \includegraphics[width=\textwidth]{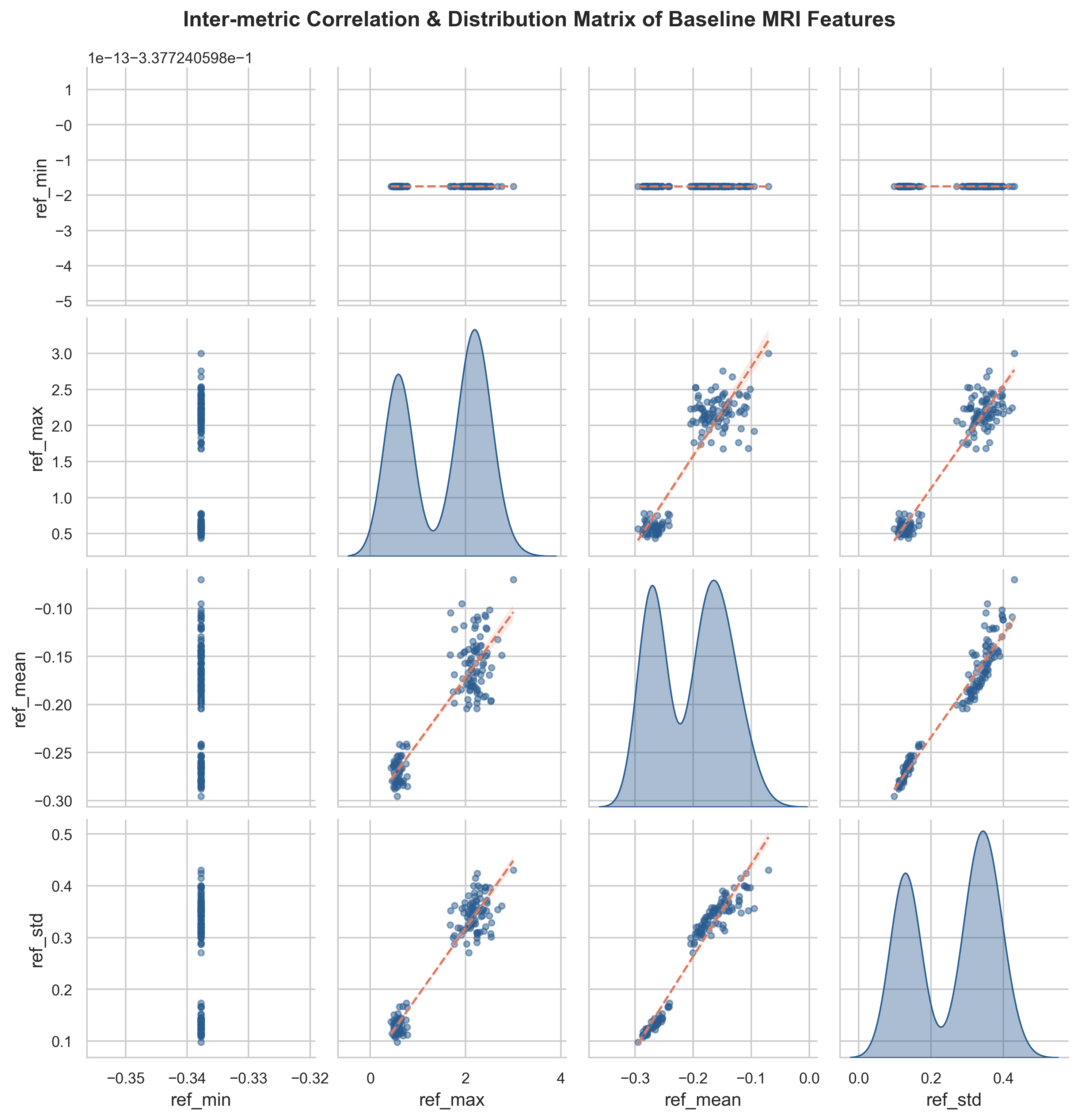}
    \caption{\textbf{Inter-metric bivariate correlation matrix of baseline imaging features.} Diagonal elements show marginal kernel density estimates. Off-diagonal panels present cross-parametric scatter plots with ordinary least-squares regression trajectories and 95\% confidence intervals, revealing the strong coupling between $I_{max}$ and image standard deviation ($I_{std}$).}
    \label{fig:pairplot_matrix}
\end{figure}

Analysis of the empirical distributions reveals several characteristic properties:
\begin{itemize}
    \item \textbf{Boundary Convergence ($I_{min}$):} Tracking the absolute lower bound verifies the consistency of background clamping strategies. In our cohort, minimum intensities converged tightly at a fixed value of $-0.338$ ($\sigma \approx 0$), confirming uniform background normalization following automated brain extraction\cite{smith2002fast} and intensity clamping\cite{shinohara2014volumetric}. This is consistent with findings by Gao et al.\cite{gao2019universal}, who observed reproducible intensity baselines across multi-center MRI acquisitions following standardized preprocessing.

    \item \textbf{Peak-Intensity Heterogeneity ($I_{max}$):} In contrast to the stable lower bounds, maximum intensities display a heavy-tailed distribution\cite{ghazvanchahi2024effect}, ranging from $0.518$ to $3.000$ in our cohort. This variance reflects localized high-contrast entities---such as contrast-enhanced vascular structures or hyperintense pathology---that appear only in specific sub-populations, and is strongly associated with domain shift in multi-scanner MRI studies\cite{ghazvanchahi2024effect}.

    \item \textbf{Bivariate Feature Coupling:} Joint-distribution analysis reveals a pronounced linear collinearity between $I_{max}$ and $I_{std}$: as peak signal intensity increases, global contrast scales proportionally, while the tissue-level mean ($I_{mean}$) remains stable. Such intensity-driven feature couplings are well documented in multi-parametric MRI radiomics and image biomarker standardization research\cite{zwanenburg2020image,gillies2016radiomics}.
\end{itemize}

\subsection{Pillar 3: Longitudinal Temporal Trajectory Mapping}\label{sec:pillar3}

When modeling continuous pathological processes from discrete sequential observations, the elapsed temporal interval ($\Delta t$) acts as an important latent conditioning variable. Because clinical imaging events are rarely recorded at equidistant intervals, treating sequential scans as uniform timesteps introduces substantial temporal distortion into continuous-time architectures\cite{disch2025temporal, wang2025conditional}.

Recent advances in longitudinal medical imaging have underscored the importance of explicit temporal modeling. Disch et al.\cite{disch2025temporal} introduced Temporal Flow Matching (TFM), a generative trajectory method that natively supports irregular sampling intervals in 4D longitudinal imaging. Wang et al.\cite{wang2025conditional} proposed Conditional Neural ODE (CNODE) for forecasting longitudinal disease progression from irregularly sampled MRI, noting that ``traditional deep learning methods such as RNNs and transformer-based architectures typically assume regularly sampled data, making them poorly suited for handling inherently continuous data sampled with varying intervals.'' This limitation has been further formalized by Rubanova et al.\cite{rubanova2019latent} through latent ODE models explicitly designed for irregular time series, confirming that continuous-time parameterizations require rigorous interval constraints.

The third pillar maps the non-uniform distribution of temporal intervals across the cohort. Let $t_1$ denote the baseline reference observation timestamp for a given subject and $t_m$ represent the absolute timestamp of the $m$-th subsequent follow-up encounter. The elapsed interval relative to baseline is parameterized as:
\begin{equation}\label{eq:temporal}
    \Delta t_m = t_m - t_1, \quad m \in \{2, 3, \dots, M\}
\end{equation}
where $M$ denotes the variable sequence length per subject. We adopt baseline-referenced offsets rather than inter-scan deltas ($t_m - t_{m-1}$) to align with continuous-time generative models where pathological evolution is parameterized as a continuous integration from the initial presentation state.

\begin{figure}[!tbp]
    \centering
    \includegraphics[width=\textwidth]{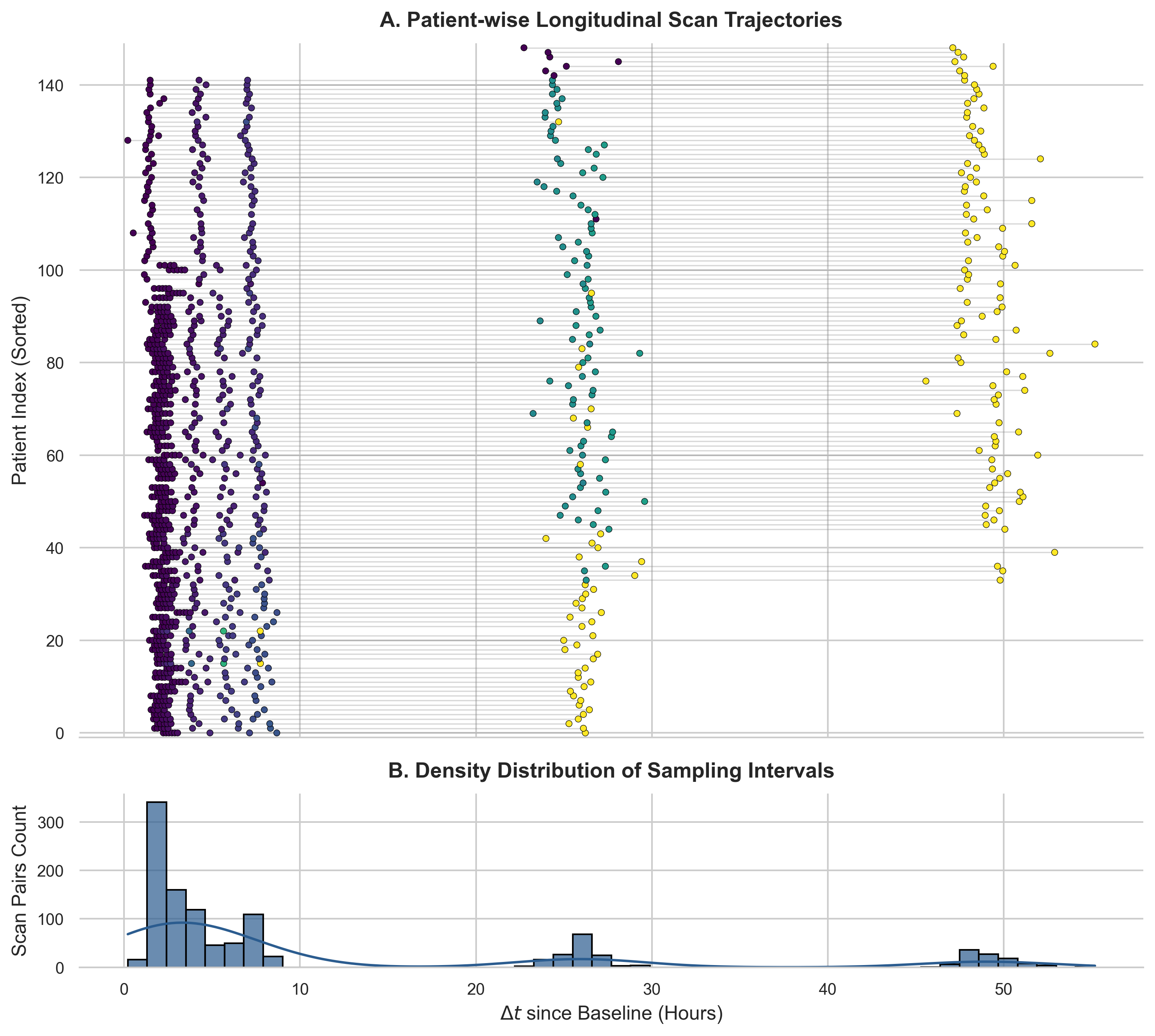}
    \caption{\textbf{Longitudinal tracking and density mapping of patient follow-up trajectories.}
    (a) Patient-wise scan trajectories sorted by total follow-up duration, colored continuously from baseline ($t=0$~h) to the long-term observation limit ($t=48$~h).
    (b) Empirical density distribution of elapsed temporal intervals ($\Delta t$), highlighting the left-skewed, non-uniform sampling characteristic of real-world clinical operations.}
    \label{fig:temporal_trajectories}
\end{figure}

In practice, clinical imaging intervals exhibit substantial irregularity driven by patient adherence, scheduling constraints, and center-specific operational factors. Rather than conforming to regularly spaced protocols, empirical follow-up patterns span a wide range of timescales---from dense acute observations in short-term dynamic monitoring to sparse, long-term assessments spanning multiple weeks or months in neuro-oncology follow-up\cite{liu2025treatment}. This non-uniformity varies systematically across patients, producing the left-skewed, variable-density temporal distributions visible in Figure~\ref{fig:temporal_trajectories}.

Characterizing these distributions is an essential mathematical prerequisite for configuring continuous-time neural architectures and provides the quantitative justification for including explicit temporal trajectory features ($\Delta t_{max}$, $M_i$) in the joint spatio-temporal clustering vector (Section~\ref{sec:stratified}), ensuring that data partitions are balanced symmetrically across both intensity and temporal dimensions independent of the underlying clinical acquisition scale.

\section{The Unsupervised Stratified Cohort Balancing SOP}\label{sec:stratified}

Given the heavy long-tailed dispersion in baseline intensity profiles and the non-uniformity of longitudinal follow-up frequencies, naive random partitioning risks introducing evaluation bias, distributional shift, and artificial performance inflation\cite{matta2024systematic,yoon2024domain}. To preserve empirical feature densities across all subsets, we replace ad hoc random grouping with an automated, deterministic, proportionate stratified sampling procedure governed by unsupervised phenotypic clustering.

\subsection{Mathematical Formulation}

To preserve the joint distribution of multi-parametric imaging attributes and longitudinal properties without manual intervention or heuristic bias, we formalize the cohort partitioning task as an adaptive, unsupervised vector-space stratification problem. Let a general clinical cohort be defined as a set of $N$ unique subjects, $\mathcal{D} = \{1, 2, \dots, N\}$. Each subject $i \in \mathcal{D}$ is mapped to a multi-parametric spatio-temporal feature vector $\mathbf{x}_i \in \mathbb{R}^D$ ($D=6$) representing their combined baseline phenotypic and longitudinal trajectory fingerprint. In our generalized framework, this feature topology is explicitly formalized as:
\begin{equation}
    \mathbf{x}_i = \left[ I_{min}^{(i)}, \, I_{max}^{(i)}, \, I_{mean}^{(i)}, \, I_{std}^{(i)}, \, \Delta t_{max}^{(i)}, \, M_i \right]^T
\end{equation}
where $I_{min}^{(i)}$, $I_{max}^{(i)}$, $I_{mean}^{(i)}$, and $I_{std}^{(i)}$ encapsulate the first and second central moments of the baseline voxel intensity distribution, while $\Delta t_{max}^{(i)}$ represents the continuous maximum temporal tracking duration, and $M_i$ captures the discrete variable sequence length unique to subject $i$'s clinical longitudinal lifecycle. Although $I_{min}$ exhibits near-zero variance in this cohort (Section~\ref{sec:pillar2}), it is retained in $\mathbf{x}_i$ for generalizability to datasets where background normalization may be incomplete.

To prevent arbitrary sub-space fragmentation or rigid, ad-hoc assumptions regarding the data structure, the optimal number of phenotypic strata $K^*$ is determined dynamically from the empirical variance of the input data. We evaluate the convergence profile of the within-cluster sum of squares ($\text{WCSS}$) across a continuous range of candidate cluster sizes $K \in [2, K_{max}]$. The parameter $\text{WCSS}(K)$ quantifies the global intra-cluster variance and is expressed as:
\begin{equation}
    \text{WCSS}(K) = \sum_{k=1}^{K} \sum_{i \in S_k} \left\| z(\mathbf{x}_i) - \bm{\mu}_k \right\|^2
\end{equation}
where $z(\cdot)$ denotes a standardizing transformation ($z$-score normalization) applied to eliminate scale disparities across heterogeneous intensity and temporal domains, $S_k$ denotes the disjoint subset of subjects assigned to the $k$-th cluster, and $\bm{\mu}_k \in \mathbb{R}^D$ represents the localized cluster centroid vector. The optimal number of strata $K^*$ is mathematically localized by identifying the maximum curvature or inflection point (the ``elbow'') of the $\text{WCSS}(K)$ function. This selection is governed by an objective, derivative-based geometric angle criterion\cite{satopaa2011finding, herdiana2025precise}, ensuring reproducibility and auditability by replacing subjective visual interpretation with analytical optimization.

Upon resolving the optimal $K^*$, the final unweighted optimization objective partitions the global cohort into $K^*$ distinct, non-overlapping phenotypic strata\cite{macqueen1967some}  $\mathbf{S}^* = \{S_1, S_2, \dots, S_{K^*}\}$ that minimize the multi-dimensional isotropic spatial variance:
\begin{equation}\label{eq:kmeans}
    \mathbf{S}^* = \arg\min_{\mathbf{S}} \sum_{k=1}^{K^*} \sum_{i \in S_k} \left\| z(\mathbf{x}_i) - \bm{\mu}_k \right\|^2
\end{equation}

\textit{Information Leakage Justification:} Because our feature mapping $\mathbf{x}_i$ extracts only unsupervised, baseline first- and second-order statistics and temporal sequence intervals -- without incorporating any downstream clinical label, pathology annotation, or prediction target -- the clustering step is entirely label-free. It performs an unsupervised population-density alignment rather than supervised predictive training, and therefore avoids the specific class of data leakage that arises from letting outcome information influence partition design.

\textit{Empirical Optimization Case Study:} This pipeline was validated on our structural longitudinal cohort of $N = 149$ baseline subjects. To reconcile the differing scales of image-contrast and longitudinal-temporal (hours) dimensions, we applied $z$-score standardization prior to clustering. The elbow-optimization procedure scanned candidate cluster counts up to $K_{max} = 8$, reducing WCSS from $246.56$ at $K=2$ to a converged floor of $45.05$ at $K=8$, with a clear geometric inflection at $K^* = 4$ (WCSS $= 99.83$). The cohort was accordingly partitioned into four spatio-temporal sub-populations of size $N_0 = 41$, $N_1 = 56$, $N_2 = 33$, and $N_3 = 19$. As shown in Table~\ref{tab:comprehensive_split_comparison}, this partition serves as the basis for subsequent proportionate stratified splitting or cross-validation, closely aligning cross-subset baseline and temporal imaging statistics.

\begin{figure}[!tbp]
    \centering
    \begin{subfigure}[b]{0.78\textwidth}
        \centering
        \includegraphics[width=0.6\textwidth]{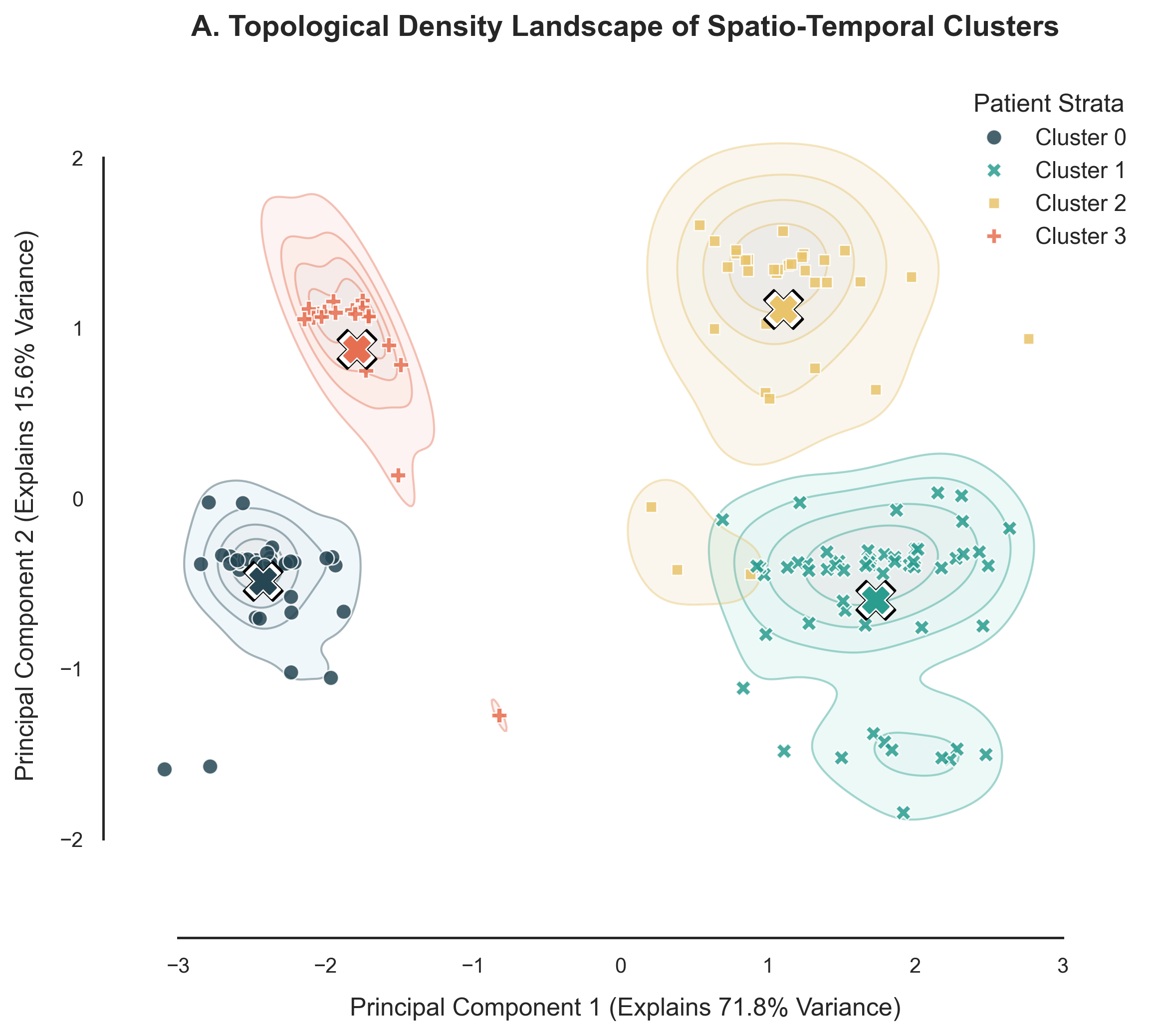}
        \caption{}
        \label{fig:sub_topological_landscape}
    \end{subfigure}

    \begin{subfigure}[b]{1.0\textwidth}
        \centering
        \includegraphics[width=\textwidth]{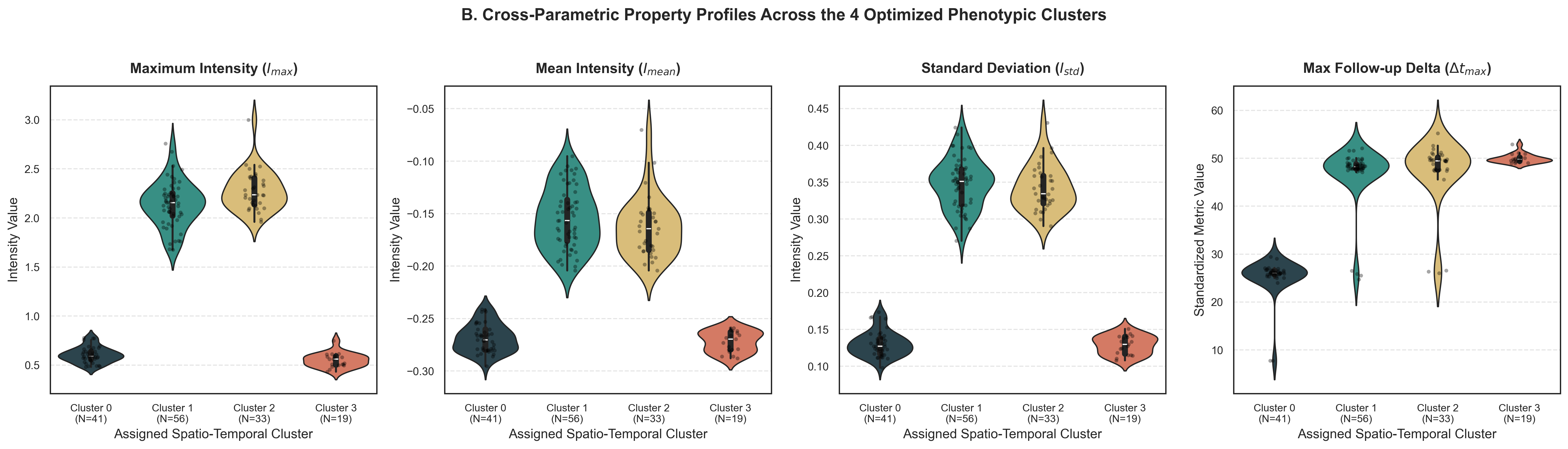}
        \caption{}
        \label{fig:sub_spatiotemporal_violins}
    \end{subfigure}

    \caption{\textbf{Unsupervised spatio-temporal cohort stratification and phenotypic profiling ($K^*=4$).}
    (a) Two-dimensional projection of the 6D joint feature space onto its leading principal components, overlaid with per-cluster kernel density estimation (KDE) contours. The resulting density fields show clear geometric separation between phenotypes, with crossed markers ($\mathbf{\times}$) indicating the corresponding cluster centroids.
    (b) Cross-parametric profiling of voxel intensity statistics ($I_{max}$, $I_{mean}$, $I_{std}$) and maximum follow-up duration ($\Delta t_{max}$). Violin and inner box plots show the empirical density per cluster with subject counts ($N$), illustrating how the pipeline achieves variance stabilization and structural alignment prior to downstream partitioning.}
    \label{fig:spatiotemporal_cohort_balancing_framework}
\end{figure}

This joint spatio-temporal projection partitions the full 149-patient cohort into $K^* = 4$ well-separated phenotypic strata. Figure~\ref{fig:spatiotemporal_cohort_balancing_framework} provides a visual validation of this data-driven alignment.

Figure~\ref{fig:sub_topological_landscape} shows the 6D joint feature representation projected onto its leading principal components. The KDE contours reveal smooth, well-separated density boundaries between clusters, and the crossed markers ($\mathbf{\times}$) mark the centroids computed by Equation~\ref{eq:kmeans}, indicating that the optimization converges to stable, high-density centers without fragmentation.

To characterize the phenotypic and longitudinal profile of each resolved stratum, Figure~\ref{fig:sub_spatiotemporal_violins} presents a cross-parametric feature grid spanning intensity and temporal characteristics across all four strata. The four sub-populations show clear between-cluster heterogeneity alongside internal consistency: Cluster 3, for instance, captures a sub-population with elevated maximum intensity ($I_{max}$) and longer follow-up ($\Delta t_{max}$), whereas Cluster 1 corresponds to a stable, low-contrast core cohort. Enforcing the proportionate stratified SOP across these four strata balances these skewed statistics across the Train, Validation, and Test subsets, mitigating the risk of out-of-distribution evaluation.

\subsection{Deterministic Partitioning Protocol and Multi-fold Generalization}

Once subjects are assigned to clusters, a deterministic allocation protocol is applied. The framework is agnostic to the specific partitioning configuration and natively supports arbitrary user-defined split ratios (e.g., the 70\%:15\%:15\% split used in this study) as well as stratified $M$-fold cross-validation. Rather than shuffling the full cohort, the procedure operates intra-cluster, performing proportionate stratified sampling within each phenotypic group. This ensures that:
\begin{itemize}
    \item Every feature sub-space---including rare sub-populations with low sample counts---is proportionally represented across all training and evaluation boundaries, preventing any subset from being deprived of specific imaging profiles.
    \item The empirical data density and temporal information entropy (total sequential pairs and elapsed intervals) remain consistent across partitions, mitigating the risk of out-of-distribution evaluation failures or artificially inflated validation performance.
\end{itemize}

The limitations of conventional partition strategies become directly measurable when the resulting subsets are compared. Under naive random shuffling, the splitting process is blind to inter-subject feature distributions. Because sample sizes in medical imaging cohorts are typically small, purely stochastic assignment has no mechanism to guarantee demographic, clinical, or covariate homogeneity across splits; downstream frameworks trained on such splits therefore remain vulnerable to unacknowledged, hidden stratification and distribution shift between training and evaluation.

Table~\ref{tab:comprehensive_split_comparison} quantifies this volatility. Under naive random shuffling, the validation and test subsets are disproportionately populated with hyperintense outlier cases, yielding elevated maximum-intensity averages ($Val\ I_{max} = 1.884$, $Test\ I_{max} = 1.878$) relative to the under-represented training subset ($1.404$). This asymmetry corresponds to a cross-subset intensity misalignment exceeding 34.1\%, effectively simulating an artificial dataset shift within the evaluation pipeline. The random split also skews the temporal domain, pushing the test-subset mean tracking interval up to $11.24$~hours.

In contrast, our spatio-temporal stratified SOP substantially mitigates these sampling anomalies. As shown in the upper tier of Table~\ref{tab:comprehensive_split_comparison}, the data-driven balancing protocol confines cross-subset variation in the multi-parametric intensity metrics to under 2.1\% (Train $I_{max} = 1.544$ vs.\ Test $I_{max} = 1.531$). The temporal trajectories are similarly regularized, with elapsed intervals ($\Delta t$) aligning closely around the population mean ($\sim 10.5$~hours) across the Train, Validation, and Test subsets. By mapping patient profiles into a four-cluster phenotype space ($K^*=4$) prior to splitting, the framework distributes the underlying imaging characteristics with close statistical equivalence across subsets, providing a more consistent optimization environment for downstream models.

\begin{table}[!tbp]
\centering
\small
\caption{\textbf{Quantitative cross-subset alignment comparison between the proposed spatio-temporal stratified protocol and conventional naive random shuffling baseline ($K^*=4$).}}
\label{tab:comprehensive_split_comparison}

\resizebox{\textwidth}{!}{%
\begin{tabular}{llcccc}
\toprule
\textbf{Splitting Framework} & \textbf{Cohort Subset} & \textbf{Patient Count} & \textbf{Mean $M_{i} \pm \text{SD}$} & \textbf{Mean $I_{max} \pm \text{SD}$} & \textbf{Mean $\Delta t$ (Hours)} \\
\midrule
\textbf{Proposed Protocol} & Train Set & 105 & $7.38 \pm 2.49$  & $1.544 \pm 0.81$ & $10.76 \pm 14.5$ \\
\textit{(Spatio-Temporal Stratified)} & Val Set   & 22  & $7.55 \pm 2.44$  & $1.563 \pm 0.85$ & $9.98 \pm 13.9$  \\
\textit{Our Method}                 & Test Set  & 22  & $7.59 \pm 2.56$  & $1.531 \pm 0.81$ & $10.97 \pm 14.6$ \\
\midrule
\textbf{Conventional Baseline} & Train Set & 105 & $7.50 \pm 2.44$ & $1.404 \pm 0.82$ & $10.42 \pm 14.1$ \\
\textit{(Naive Random Shuffling)}      & Val Set   & 22  & $7.23 \pm 2.56$ & $1.884 \pm 0.73$ & $11.36 \pm 14.9$ \\
                                       & Test Set  & 22  & $7.32 \pm 2.66$ & $1.878 \pm 0.67$ & $11.24 \pm 15.2$ \\
\bottomrule
\end{tabular}%
}
\end{table}

\subsection{Methodological Robustness and Multi-Dimensional Stress Testing}\label{sec:robustness}

To assess whether our spatio-temporal balancing protocol constitutes a generalizable statistical solution rather than an artifact of one particular split, we subject the pipeline to two complementary stress tests evaluating alignment stability across varying partition scales and across iterative cross-validation. In both settings, partition error is quantified as the absolute deviation of a subset's empirical feature mean from the full cohort mean,
\begin{equation}\label{eq:delta}
    \Delta = \left\| \bm{\mu}_{\text{subset}} - \bm{\mu}_{\text{population}} \right\|,
\end{equation}
measured independently along the peak spatial intensity axis ($I_{max}$) and the temporal follow-up interval ($\Delta t$).

\paragraph{Multi-Ratio Holdout Stress Testing.}
We first evaluate alignment performance across three distinct clinical splitting configurations: $70\%:15\%:15\%$ (baseline configuration), $80\%:10\%:10\%$ (restricted evaluation boundaries), and $60\%:20\%:20\%$ (expanded validation footprints). Each structural ratio is cross-examined under ten independently sampled randomized initialization seeds ($\mathcal{S} \in \{12, 42, \dots, 9999\}$), yielding 30 independent holdout validation environments per splitting framework.

\begin{figure}[!tbp]
    \centering
    \includegraphics[width=\textwidth]{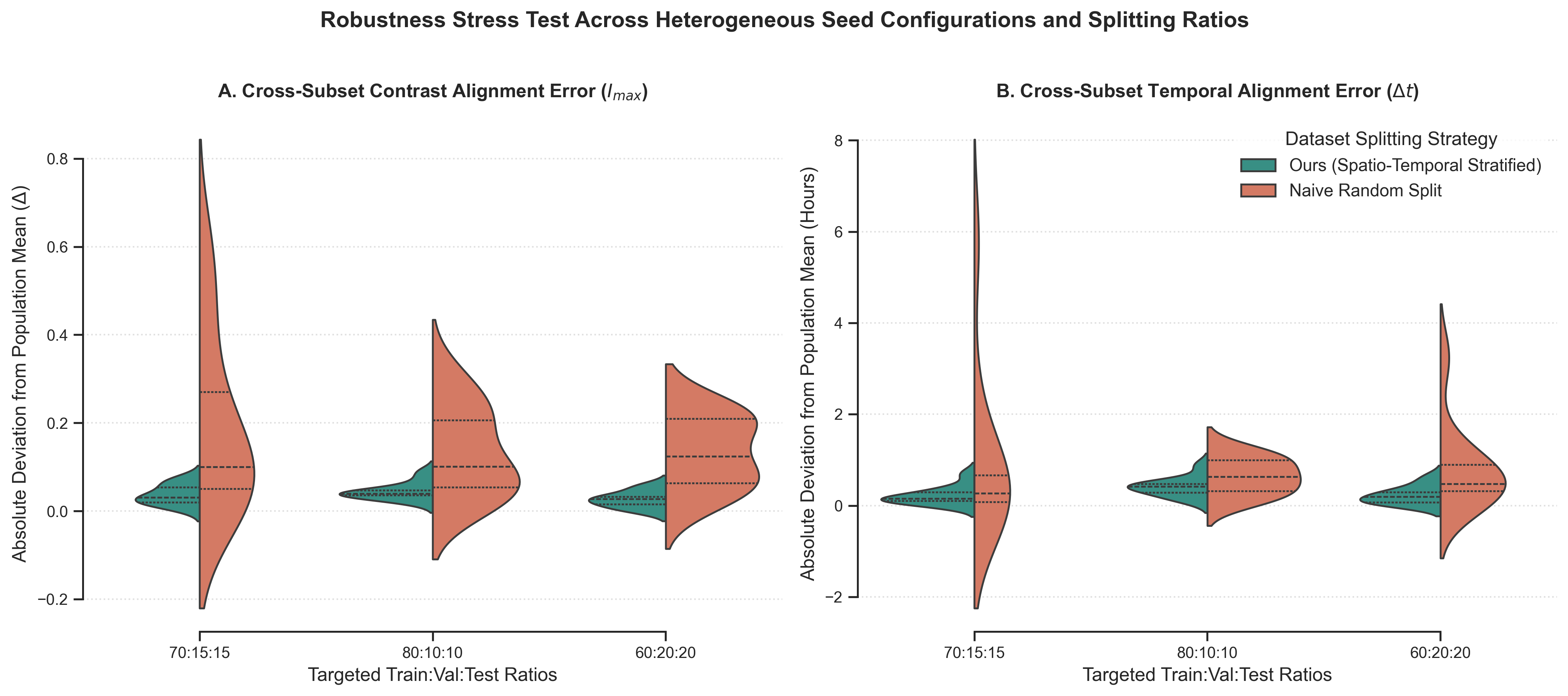}
    \caption{\textbf{Robustness stress test A: Absolute alignment deviations across diverse holdout split ratios.} Split-violin plots cross-examine the deviation distributions of the proposed spatio-temporal stratified protocol (deep cyan) against conventional naive random shuffling (orange-red). (a) Spatial intensity deviations ($I_{max}$), where naive random splitting induces heavy-tailed error distributions. (b) Temporal tracking deviations ($\Delta t$), mapping the bounded stability of the data-driven framework across variable splitting scales.}
    \label{fig:ratio_robustness_benchmark}
\end{figure}
\begin{figure}[!tbp]
    \centering
    \includegraphics[width=\textwidth]{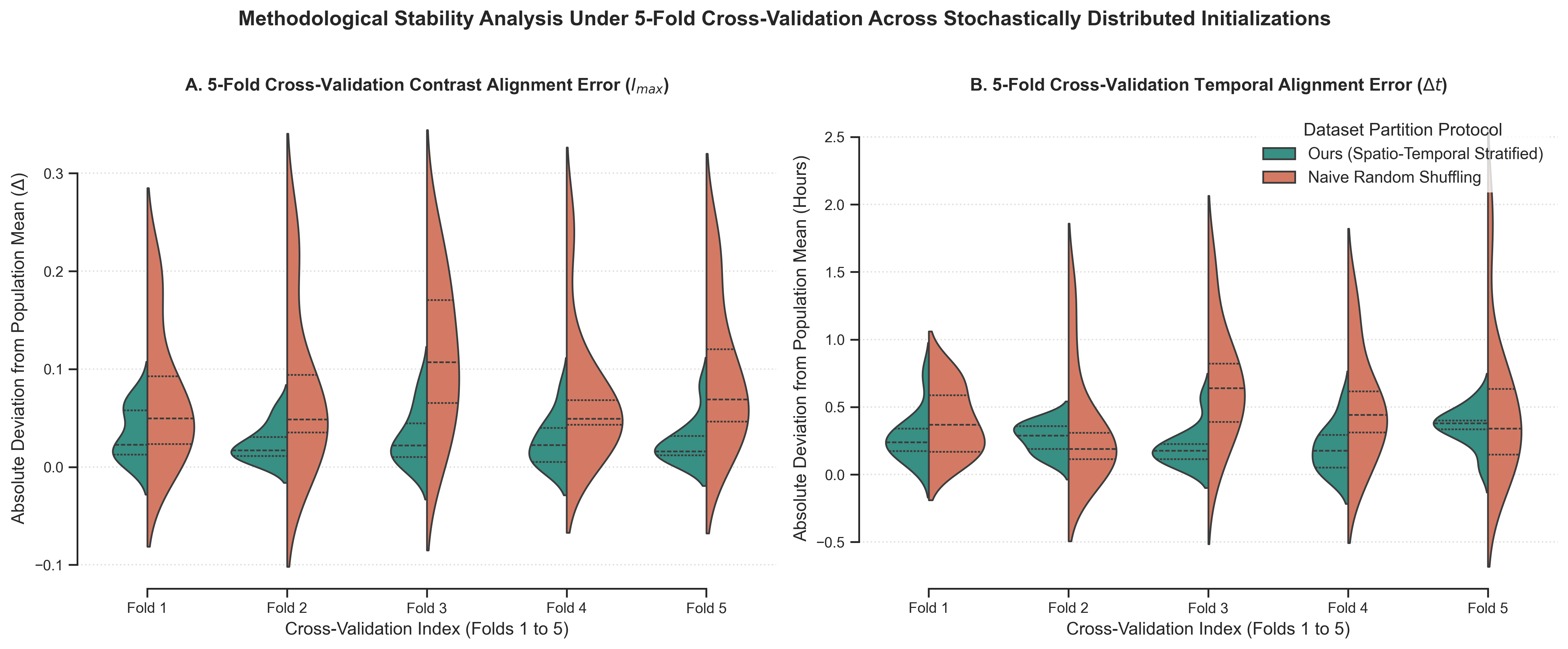}
    \caption{\textbf{Robustness stress test B: Absolute alignment deviations across iterative cross-validation folds.} Split-violin plots compare the proposed stratified protocol (deep cyan) against conventional naive random 5-fold shuffling (orange-red) across Folds 1 to 5. (a) Intensity deviations ($I_{max}$), showing severe variance dispersion and outlier propagation under naive random shuffling within Folds 2 and 5. (b) Temporal deviations ($\Delta t$), highlighting unmitigated entropy inflation breaching 4.5 hours under random shuffling within Folds 3 and 5, whereas the stratified protocol maintains tightly constrained error bounds.}
    \label{fig:5fold_robustness_benchmark}
\end{figure}

\paragraph{Stratified 5-Fold Cross-Validation Cross-Examination.}
To evaluate the protocol's cyclic invariant stability, we additionally subject the cohort to a 5-fold cross-validation regime. The 5-fold partition is executed intra-cluster to ensure that the empirical density of all four resolved phenotypic strata ($K^*=4$) remains structurally equivalent across every evaluation fold. Combined with the identical ten randomized initialization seeds, this framework generates 50 independent validation folds per method, presenting a rigorous multi-sample challenge to partition homogeneity.

\paragraph{Holistic Stability and Cross-Examination Assessment.}
Across all tested configurations, splitting ratios, and randomized seed initializations illustrated in Figure~\ref{fig:ratio_robustness_benchmark} and Figure~\ref{fig:5fold_robustness_benchmark}, the proposed spatio-temporal stratified protocol maintains a consistently tight error envelope. The localized median intensity deviations are strictly bounded within $\Delta < 0.04$ across all holdout scales and cross-validation increments, while the corresponding temporal tracking offsets remain stabilized below $\Delta < 0.5$~hours.

By contrast, the conventional label-blind ``shuffle-and-train'' baseline exhibits substantial stochastic variability and pronounced localized distribution shifts. Under the restricted $70:15:15$ holdout scale, as well as within cross-validation Folds 2 and 5, the random-partitioning intensity error reaches $\Delta = 0.45$. Temporal tracking discrepancies under random shuffling similarly exceed 4.5 hours within Folds 3 and 5. These results indicate that the proposed SOP provides reliable, seed-invariant, and scale-invariant alignment across both holdout and cross-validation regimes, whereas conventional random partitioning cannot ensure a consistent evaluation boundary across varying split scales or fold structures.

\section{Discussion}\label{sec:discussion}

\subsection{Implications for Longitudinal Deep Learning and Evaluation Integrity}

The joint spatio-temporal cohort engineering framework formalized here provides a useful statistical foundation for continuous-time sequential models and generative trajectory networks. Architectures specializing in variable-length longitudinal modeling---such as recurrent neural networks, neural ordinary differential equations (Neural ODEs), and flow-matching networks---rely on the assumption that training gradients reflect genuine underlying pathological progression rather than discrete sampling artifacts. When dataset partitioning is governed by naive random shuffling, the resulting 34.1\% cross-subset contrast misalignment forces the temporal and spatial parameter layers to partly optimize against artificial scaling variance, constraining the network's capacity to learn temporally coherent representations of disease dynamics.

Our findings also help explain the non-monotonic performance profiles often observed during longitudinal evaluation. In small clinical cohorts, evaluation metrics tend to fluctuate noticeably across intermediate follow-up intervals, largely because transitional scan sessions are clinically scarce. Standard image- or time-domain augmentations, routinely applied to inflate the number of sequential sample pairs, operate at the sample level and cannot compensate for subset-level skewness. Applying our unsupervised stratified protocol before any such augmentation ensures that irregular follow-up lifecycles are distributed symmetrically across validation boundaries. This helps insulate downstream architectures from artificial validation inflation, so that reported structural-fidelity and signal-to-noise metrics are more likely to reflect genuine model generalizability rather than a favorable random split.

\subsection{Relation to Domain Shift Mitigation}

Our cohort engineering methodology is best understood as a proactive form of domain shift mitigation. Rather than adapting models post hoc to a distributional shift already present in the data\cite{guan2022domain}, the SOP aims to prevent that shift from being introduced during dataset construction in the first place. This is consistent with Matta et al.\cite{matta2024systematic}, who note that existing domain generalization methods still lack adequate evaluation protocols and benchmarks -- partitioning quality is one such protocol-level gap.

The intensity-standardization literature provides complementary context: MRI normalization methods, including $z$-score normalization, WhiteStripe\cite{shinohara2014standardized}, Nyul--Udupa histogram matching\cite{nyul1999standardizing}, and GAN-based harmonization\cite{gao2019universal}, have been shown to improve cross-scanner generalization\cite{ghazvanchahi2024effect}. Our statistical fingerprinting pillar complements these methods rather than replacing them: it provides a quantitative audit trail that verifies normalization effectiveness and flags residual heterogeneity prior to training.

\subsection{Temporal Considerations}

The non-uniform sampling distribution documented in Figure~\ref{fig:temporal_trajectories} has direct implications for continuous-time deep learning architectures. Tak et al.\cite{tak2025longitudinal} showed that temporal learning models for longitudinal MRI benefit incrementally from additional historical scans, with performance typically plateauing between three and six scans -- but these gains assume that training and validation distributions are temporally matched, an assumption that random partitioning is not guaranteed to satisfy.

Zhao et al.\cite{zhao2021longitudinal} proposed Longitudinal Self-Supervised Learning (LSSL), which disentangles brain-age factors through temporal-order supervision across multiple MRIs per subject. Approaches of this kind depend on the training set's temporal dynamics being representative of the full cohort; our temporal trajectory mapping (Pillar 3) provides the distributional characterization needed to verify that this condition holds.

\subsection{Limitations and Future Directions}

This work has several limitations. First, the current implementation assumes that the baseline fingerprint $\bm{\chi}_i$ captures the dominant sources of inter-subject variation. In multi-modal settings or cohorts with more complex covariate structure, additional features -- such as demographic variables or clinical scores -- may need to be incorporated into the clustering space\cite{yoon2024domain}. Balancing a cohort on a fixed set of summary features can also, in principle, under-represent rare sub-populations that a low-dimensional fingerprint fails to separate from the bulk distribution; we treat this as an open trade-off that warrants dedicated evaluation in future deployments rather than one resolved by the present study.

Second, although the clustering configuration in this case study ($K^* = 4$) was determined dynamically via the elbow method, the optimal number of strata will vary with cohort size, dimensionality, and imaging modality. We treat $K^*$ as an empirical function of the cohort's internal dispersion rather than a fixed hyperparameter, but the elbow heuristic itself is only one of several reasonable choices. Future work could incorporate complementary validation heuristics -- such as the gap statistic\cite{tibshirani2001estimating} or silhouette analysis -- to further automate, or cross-check, stratum selection in larger or more heterogeneous cohorts.

Third, our validation is based on a single-center, single-modality cohort ($N=149$); while the underlying feature-space formalism (Section~\ref{sec:stratified}) is not tied to this modality, its behavior on multi-center or cross-modality cohorts, where the dominant sources of covariate shift may differ, remains to be tested empirically.

Finally, while the SOP enforces distributional homogeneity at the partitioning stage, it does not address domain shift that emerges later, during model deployment. Integration with test-time adaptation\cite{guan2022domain} or continual learning\cite{yoon2024domain} strategies is a natural direction for future work. More broadly, whether stratified partitioning of this kind translates into a measurable downstream evaluation benefit is itself dataset-dependent: it hinges on whether cross-subset covariate imbalance, rather than case-level outcome variance, is the dominant driver of partition sensitivity for a given task and cohort. We view this as an open empirical question to be assessed per dataset, rather than a property guaranteed by the partitioning procedure itself.

\section{Conclusion and Future Guidelines}\label{sec:conclusion}

This study formalizes a repeatable methodology for cohort curation and dataset partitioning in longitudinal medical imaging. By replacing arbitrary random partitioning with multi-parametric statistical stratification, we obtain a more stable validation boundary for deep learning workflows. Our empirical validation on a 149-patient longitudinal, contrast-enhanced $T1$-weighted brain MRI cohort shows that the proposed SOP substantively reduces the maximum cross-subset feature bias from over 34.1\% to under 2.1\%, using the alignment deviation $\Delta$ defined in Equation~\ref{eq:delta}.

\textbf{Recommendations for future cohort engineering:}
\begin{enumerate}
    \item Do not pass raw clinical sequential data directly into a training pipeline without first running a spatial coordinate and grid-spacing audit (Equation~\ref{eq:spatial}).
    \item Perform multi-parametric statistical fingerprinting and cross-feature density profiling (Figure~\ref{fig:intensity_distributions}) to identify heavy-tailed pathology clusters or intensity outliers before partitioning.
    \item Replace naive random partitioning with an elbow-optimized spatio-temporal stratified protocol (Equation~\ref{eq:kmeans}) to improve structural and statistical reproducibility across training and evaluation partitions.
    \item Restrict subsequent data augmentation or stochastic domain perturbation to operate downstream of the cohort stratification step, applied only within the training subset.
\end{enumerate}

Following these recommendations can improve the statistical reliability and generalizability of longitudinal medical deep learning models, moving data preparation from an ad hoc step toward an auditable, reproducible stage of the modeling pipeline -- a useful prerequisite for building clinically trustworthy AI systems.

%

\bibliography{references}

\end{document}